\documentclass[conference]{IEEEtran}
\IEEEoverridecommandlockouts
\usepackage{cite}
\usepackage{amsmath,amssymb,amsfonts}
\usepackage{algorithmic}
\usepackage{graphicx}
\usepackage{textcomp}
\usepackage{xcolor}
\def\BibTeX{{\rm B\kern-.05em{\sc i\kern-.025em b}\kern-.08em
    T\kern-.1667em\lower.7ex\hbox{E}\kern-.125emX}}

\usepackage[caption=false]{subfig}
\usepackage{svg}
\usepackage{url}
\usepackage{multicol}
\usepackage{multirow}
\usepackage{color}
\usepackage{hhline}
\usepackage{pifont}
\usepackage{comment}
\usepackage{cleveref}
\usepackage{tablefootnote}
\usepackage{flushend}
\usepackage[ruled,vlined]{algorithm2e}
\definecolor{mynicegreen}{RGB}{11,102,35}
\usepackage[group-separator={,}]{siunitx}

\newcommand{\model}{{\textsc{HHGNN-CHAR}}}

 \newcommand{\squishlist}{
	\begin{list}{$\bullet$}
		{ \setlength{\itemsep}{0pt}
			\setlength{\parsep}{3pt}
			\setlength{\topsep}{3pt}
			\setlength{\partopsep}{0pt}
			\setlength{\leftmargin}{1.5em}
			\setlength{\labelwidth}{1em}
			\setlength{\labelsep}{0.5em} } }
	
	\newcommand{\squishlisttwo}{
		\begin{list}{$\bullet$}
			{ \setlength{\itemsep}{0pt}
				\setlength{\parsep}{0pt}
				\setlength{\topsep}{0pt}
				\setlength{\partopsep}{0pt}
				\setlength{\leftmargin}{2em}
				\setlength{\labelwidth}{1.5em}
				\setlength{\labelsep}{0.5em} } }
		
		\newcommand{\squishend}{
	\end{list}}
	
\begin{document}

\title{Heterogeneous Hyper-Graph Neural Networks for Context-aware Human Activity Recognition}

\author{\IEEEauthorblockN{%
Wen Ge\textsuperscript{\textsection},
Guanyi Mou\textsuperscript{\textsection},
Emmanuel O. Agu,
Kyumin Lee}
\IEEEauthorblockA{\textit{Computer Science Department}, 
\textit{Worcester Polytechnic Institute}
Worcester, MA, USA \\
{\{wge, gmou, emmanuel, kmlee\}}@wpi.edu}
}

\maketitle
\begingroup\renewcommand\thefootnote{\textsection}
\footnotetext{Equal contribution}
\endgroup

\begin{abstract}
Context-aware Human Activity Recognition (CHAR) is challenging due to the need to recognize the user's current activity from signals that vary significantly with contextual factors such as phone placements and the varied styles with which different users perform the same activity. In this paper, we argue that context-aware activity visit patterns in realistic in-the-wild data can equivocally be considered as a general graph representation learning task. We posit that exploiting underlying graphical patterns in CHAR data can improve CHAR task performance and representation learning. Building on the intuition that certain activities are frequently performed with the phone placed in certain positions, we focus on the  \textit{context-aware human activity} problem of recognizing the $<$Activity, Phone Placement$>$ tuple. We demonstrate that CHAR data has an underlying graph structure that can be viewed as a heterogenous hypergraph that has multiple types of nodes and hyperedges (an edge connecting more than two nodes).
Subsequently, learning $<$Activity, Phone Placement$>$ representations becomes a graph node representation learning problem. After task transformation, we further propose a novel Heterogeneous HyperGraph Neural Network architecture for Context-aware Human Activity Recognition (HHGNN-CHAR), with three types of heterogeneous nodes (user, phone placement, and activity). Connections between all types of nodes are represented by hyperedges. Rigorous evaluation demonstrated that on an unscripted, in-the-wild CHAR dataset, our proposed framework significantly outperforms state-of-the-art (SOTA) baselines including CHAR models that do not exploit graphs, and GNN variants that do not incorporate heterogeneous nodes or hyperedges  with overall improvements
14.04\% on Matthews Correlation Coefficient (MCC) and 
7.01\% on Macro F1 scores. 
\end{abstract}


\section{Introduction}
\label{sec:intro}

\textbf{Motivation: } Context-aware Human Activity Recognition  (CHAR)~\cite{rault2017survey} tries to recognize the user's activity while being aware of their current context. CHAR is an important task in Context-aware (CA)~\cite{lee2007deploying} Systems, targeting diverse real-life problems~\cite{chen2018evaluating, lindqvist2011undistracted}.
To generate realistic, labeled CHAR datasets for supervised learning, apps installed on the smartphones (or smartwatches) of human volunteers\footnote{We use ``volunteers'', ``participants'' and ``users'' interchangeably}, continuously gather sensor data (e.g., accelerometer) as they live their lives while performing various activities. Users are also prompted periodically to provide ground truth context-aware activity labels. While many definitions of user context exist in the literature~\cite{abowd1999towards,dey2001conceptual}, 
in this paper, since certain activities are frequently performed with the phone placed in certain positions, we focus on the specific \textit{CHAR} problem of recognizing the $<$Activity, Phone Placement$>$ tuple. The overarching goal of this paper is to create a pattern recognition model that can simultaneously infer a user's current activity and phone placement from sensor data gathered from their smartphone. We focus on CHAR on smartphones, which are now nearly ubiquitously owned.

\textbf{Prior Work}: has explored improving Human Activity Recognition (HAR) by exploiting underlying graphical structures in activity data. Martin et al.~\cite{martin2018graph} modeled human mobility patterns by using GPS coordinates to build multiple personalized graphs, and then predicted target HAR labels using two Graph Convolution Networks (GCNs)~\cite{kipf2016semi}. 
Mohamed et al.~\cite{mohamed2022har} proposed HAR-GCNN, which exploited correlations between chronologically adjacent sensor measurements to predict missing activity labels. Selected activities were used to build a fully connected graph where each node contained sensors values and corresponding activitly labels. A GCN was employed to learn node embeddings, followed by a sequence of CNNs to predict activity labels. 
One limitation of HAR-GCNN is that it is incapable of making predictions for nodes with completely missing labels as is common in in-the-wild HAR datasets. 
\par
\smallskip \textbf{Novelty: } Our work differs from prior work in two ways: 
 
\smallskip\noindent \textbf{1) The CHAR task was not explored. } Prior work~\cite{martin2018graph,mohamed2022har} recognized activities without factoring in phone placement or inter-user activity performance style, which results in an oversimplification of the problem. Such HAR models would underperform when deployed in the wild~\cite{suh2022adversarial,attal2015physical, berchtold2010extensible}. 
    
\smallskip\noindent\textbf{2) Graphs were built using data from specific sensors and labels.} Some sensor data can be privacy sensitive (e.g., GPS),  and less than 50\% of users are willing to grant permission access permission~\cite{dogrucu2020moodable}. In contrast, our approach derives the graph directly from CHAR data based solely on label co-occurrence information observed in the training set.

\begin{figure}[t]
\centering
    \includegraphics[width=.45\linewidth]{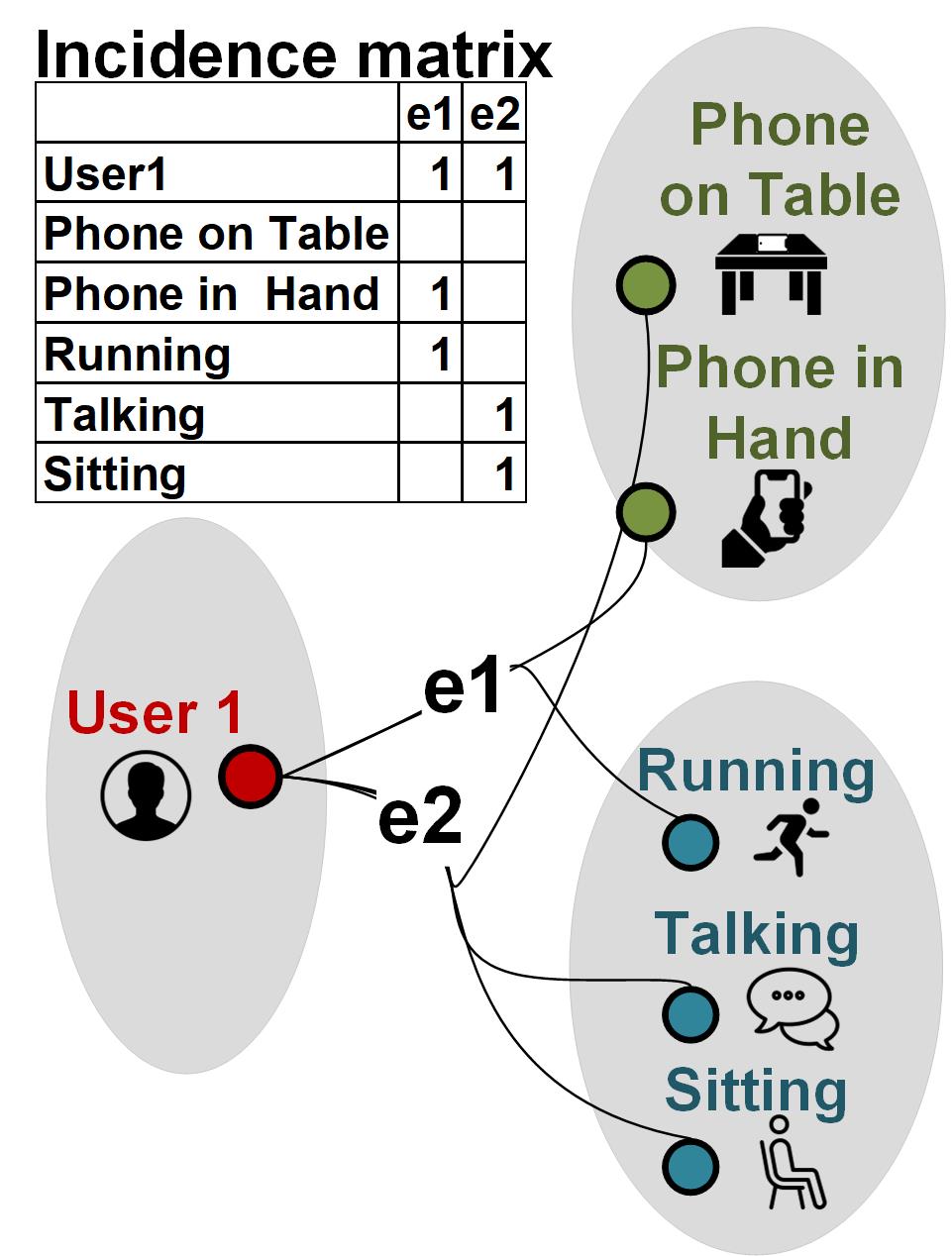}
    \vspace{-5pt}
    \caption{Sample CHAR graph that has heterogeneous nodes and hyperedges}
    \label{fig:graph}
    \vspace{-15pt}
\end{figure}

\textbf{Our approach:} 
We propose a generalizable GNN-based approach that improves CHAR performance by exploiting graphical patterns and internal relationships between different entities in CHAR data without requiring external information. We propose 1) \textit{A CHAR graph: } that has three types of graph entity/nodes (activities, phone placements, and users) and edges as the aggregated mean feature values of instances that have similar corresponding labels connecting them.  
2) A method of encoding 
CHAR data,  transforming the associated recognition task into the equivalent heterogeneous hypergraph representation learning problem that is inspired by user-item interaction graphs commonly utilized in recommender systems~\cite{he2020lightgcn,wang2019neural,zhu2016heterogeneous}, and 3) A novel deep heterogenous hypergraph Graph Neural Network (GNN)-based learning model to solve the CHAR task. 


The obtained CHAR graphical representation is heterogeneous~\cite{zhang2019heterogeneous} because the activity, phone placement, and user attributes that we define as nodes in our graph, are actually different types of nodes.
As participants may perform multiple activities with specific phone placement concurrently, connecting all activities performed by a given user to valid phone placements results in hyperedges~\cite{feng2019hypergraph}.
Such a one-to-many mapping condition is formulated as a multi-label problem in which phone placement and activity labels may co-occur. The example CHAR graph shown in Fig.~\ref{fig:graph} has two hyperedges: $e1$ and $e2$, where hyperedge \textit{e2} represents the User1 putting the Phone On Table while Sitting and Talking at the same time.
After graphical encoding and transformation, a CHAR \textbf{H}eterogeneous \textbf{H}yper\textbf{G}raph is obtained. 
In the original CHAR task, given an input signal representation, a CHAR model predicts associated context-aware activity labels. In the transformed task, given input data as a new hyperedge and the heterogeneous hypergraph discovered from the CHAR training set, a graph representation learning model can predict the labels as the most likely connecting nodes. Transforming the CHAR task into a graph representation learning task, facilitates explicit identification and exploitation of graphical relationships among three types of nodes (activity, phone placement, and user) and with the corresponding sensor data.

\textbf{Our contributions: } in this paper are as follows: 
\squishlist
\item A method for graphical encoding of CHAR data is proposed, defining three types of CHAR nodes (activity, phone placement, and user) and the corresponding signal representations as edges based on underlying relationships.

\item Encoding the CHAR task into a graph representation learning problem is proposed, establishing that the transformed CHAR graph has heterogeneity and hypergraph properties. 

\item A novel \textbf{H}eterogeneous \textbf{H}yper\textbf{G}raph \textbf{N}eural \textbf{N}etwork for \textbf{C}ontext-aware \textbf{H}uman \textbf{A}ctivity \textbf{R}ecognition (\model) is proposed, which solves the above-mentioned heterogeneous hypergraph representation learning problem in a supervised fashion. 

\item We rigorously evaluate our proposed {\model} and demonstrate that it outperforms SOTA baselines
by 
14.04\% on Matthews Correlation Coefficient and 
7.01\% on Macro F1 scores on a realistic, in-the-wild CHAR dataset. An extensive ablation study also revealed the non-trivial contributions made by each component of {\model} and the novel adaptations of incorporating graph heterogeneity and hypergraph properties. 

\squishend

To the best of our knowledge, our work is the first to reformulate the CHAR task as a generalized graph learning problem, and we propose a specific GNN framework that significantly outperforms other non-GNN SOTA approaches. 

\begin{figure*}[ht]
\centering
\vspace{-5pt}
    \includegraphics[width=.65\linewidth]{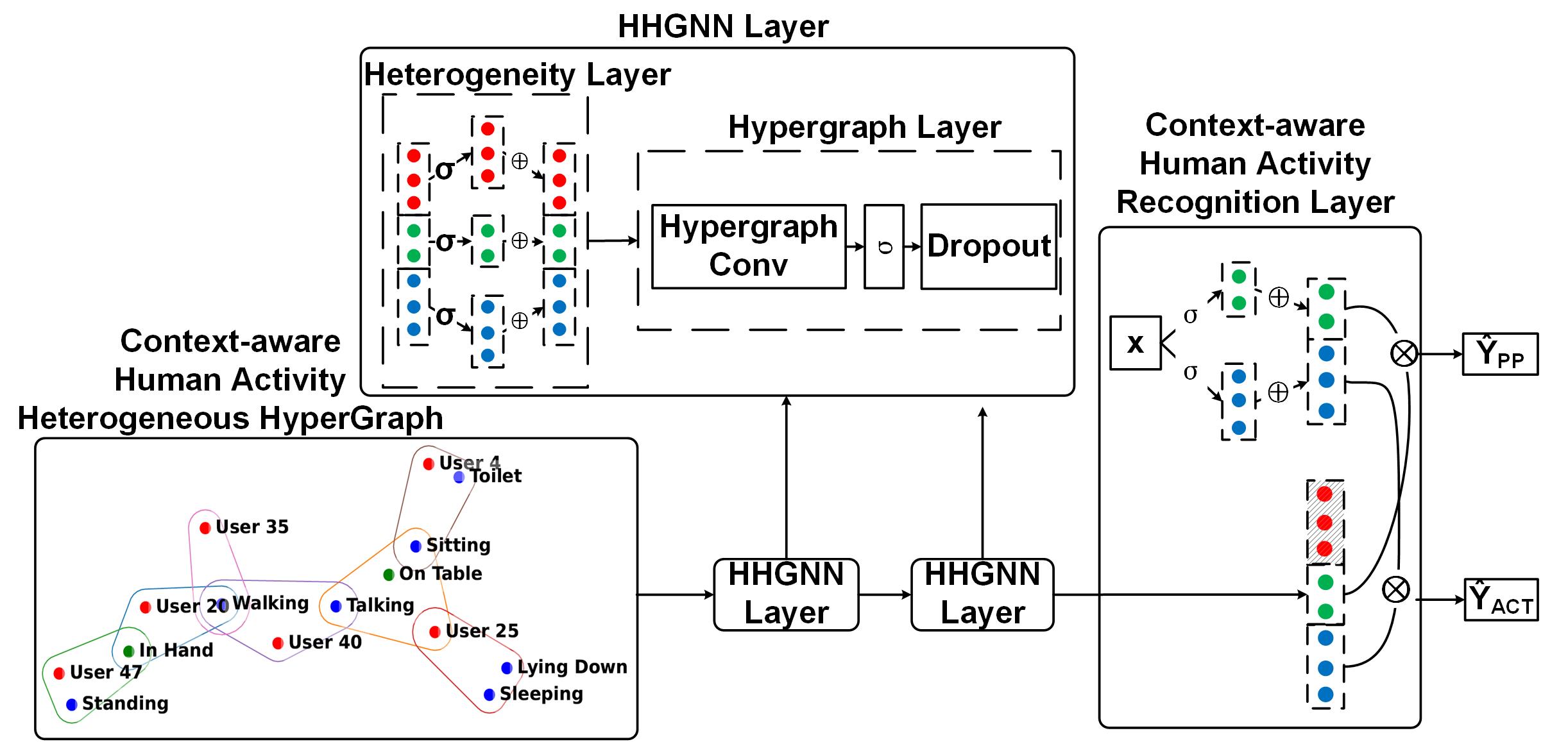}
    \vspace{-5pt}
    \caption{An Overview of our HHGNN-CHAR Framework.}
    \label{fig:framework}
    \vspace{-10pt}
\end{figure*}
\section{Proposed Framework}
%
%
\subsection{Notations and Specifications}
\label{sec:notation}
For CHAR, \textbf{Model Input} and \textbf{Graph Information} are inputs to the \textbf{Model}, which generates predictions as \textbf{Model Output} and approximations of the actual \textbf{Model Target}. At a high level, model optimization during training, model parameters are adjusted to reduce the gap between \textbf{Model Output} and the \textbf{Model Target}. In subsequent sections, unless explicitly noted, the notation conventions followed are: 
\squishlist
\item \textbf{Model Input:} The batched input data to models are denoted as $X$. Each instance within the batch is $x$. $X_{train}$ and $X_{test}$ stand for the training and testing sets, respectively. The inputs are usually features extracted from smartphone sensor signals, which often undergo pre-processing steps before being consumed by the model. 
\item \textbf{Graph Information:} Graph nodes and edges are denoted as $V$ and $E$, the 
incidence matrix (for each hyperedge, whether each node is connected) as $G_H$ (Eq.~\ref{eq:incidence}). Fig.~\ref{fig:graph} shows an example incident matrix. The hyperedge weight matrix is denoted as $G_W$ (frequency of each hyperedge), the hyperedge initial representation is $G_{attr}$, node degree matrix is $D_V=\sum_{e \in G_E} G_W(e)G_H(v,e)$ and hyperedge degree matrix is $D_E=\sum_{v\in V}G_H(v,e)$. The input $V$ for each layer in the network is $V_{in}$ while the output is $V_{out}$.
\begin{equation}
\small
    G_H(v,e) = \left\{
    \begin{aligned}
       & 0, \textit{if } v \in e, \\
       & 1, \textit{if } v \not\in e
    \end{aligned}
    \right.
    \label{eq:incidence}
\end{equation}
\item \textbf{Model Target:} Batched target labels are denoted as $Y$ and each label as lowercase $y$. Targets are the ground truth CHAR activity and phone placement labels.
\item \textbf{Model Output:} $\hat{Y}$ and $\hat{y}$ are model-generated predictions.
\item \textbf{The Model:} The general model is denoted as $M$ and the model's parameters as $\theta_{M}$.
\item \textbf{Other Specifications:} A graph is denoted as $G$, users as $U$, phone placements as $PP$, and activities as $ACT$. 
\squishend
\subsection{Graph Formation / Task Transformation}
\label{sec:graphTransform}
\noindent\textbf{The Original Task:} Given ``input data'' as smartphone sensor features $X$, performed by the simulation users $U$, an CHAR model $M$  ought to generate ``predictions'' around the $<ACT, PP>$ pairs, represented by $\hat{Y}_{ACT}, \hat{Y}_{PP}$, where $\hat{Y}_{ACT} \in R^{|ACT|}$ and $\hat{Y}_{PP} \in R^{|PP|}$. 
\begin{equation}
\small
    \hat{Y} = <\hat{Y}_{ACT}, \hat{Y}_{PP}> = \theta_{M}(X)
\end{equation}
Traditional CHAR approaches directly model $\theta_{M}$ using machine learning / deep learning techniques such as Multi-Layer Perceptrons (MLP) or LightGBM  to enable the model automatically learn the hidden correlations between $X$ and $Y$. However, these approaches lack the explicit modeling of 1) Internal relationships within $Y$, especially ${Y}_{PP}$ and ${Y}_{ACT}$; 2) User-specific information that accounts for inter-user variability in visiting various context, which ultimately impacts its performance: correlation between $U$ and $Y$.

\smallskip\noindent\textbf{The Transformed CHAR Task:} Considering the lack of explicit modeling correlations among $X, Y, U$, we transform the CHAR task into a graphical representation learning problem. Given $(X_{train}, Y_{train})$, we formulate an undirected heterogeneous hypergraph $G_{train}$ with the following relationships:
\begin{equation}
\small
\begin{split}
    G_{train} & = <V, E> \\
    V & = \{v_i | i \in (U, PP, ACT)\} \\
    E & = \{e_i | e_i \in V \times V^{*}\}
\end{split}
\end{equation}
where $*$ is noted for possible hyperconnections (connecting more than two nodes). 
Thus, a graph learning model $M_G$ consumes the graph G as input and generates the learned node representations $V_{rep}$ as output. The instance encoding model $M_{enc}$ then transform the new feature vector $x$ into the same feature dimensions as the node representation, denoted as $x_{enc}$. Intuitively, one may treat $x$ as a hyperedge with unknown connections to nodes. A third decision-making model $M_{dec}$ consumes both graph node representations and the new hyperedge information, and predicts its node connections.
\begin{equation}
\small
    \begin{split}
        V_{rep} &= M_G(G_{train}) \\
        x_{enc} &= M_{enc}(x) \\
        \hat{Y} = <\hat{Y}_{ACT}, \hat{Y}_{PP}> &= M_{dec}(V_{rep}(PP, ACT), x_{enc})
    \end{split}
\end{equation}
In the following sections, we will explain how each separate component in the deep learning model is designed.

\subsection{Graph Neural Network Design: {\model}}


\label{sec:network}
Our {\model} network (Fig.~\ref{fig:framework}),  is composed of three types of fundamental sub-layers:  Heterogeneity Layer, Hypergraph Layer, and Context-aware Human Activity Recognition Layer.
The Heterogeneity Layer takes in node representations from either the constructed/learned Context-aware Human Activity Heterogeneous HyperGraph and encodes Heterogeneity into representations by projecting different types of nodes into their own corresponding hidden spaces. The Hypergraph Layer consumes the outputs from the Heterogeneity Layer and adds  Hypergraph Convolutions plus activations and dropouts to address the hyperedges' characteristics. A combination of the Heterogeneity Layer and the Hypergraph Layer can be viewed as a HHGNN Layer block, which is stacked in order to learn the center node representations from more hops of its neighborhood. Lastly, the CHAR layer receives fine-tuned node representations from blocks of the HHGNN Layers and the task requires instance input, comparing them and deriving the possible nodes corresponding to the input instance input.

\smallskip\noindent\textbf{Heterogeneity Layer:} is introduced to explicitly address the different types of nodes within the graph (i.e., Us, PPs, and ACTs). A given $V$, the input vector to the layer, is broken into node groups, which are handled by separate linear parameters and activations, and eventually, concatenate results as $V_{out}$
\begin{equation}
\small
    \begin{aligned}
    V_{in} & = [V_{U}, V_{PP}, V_{ACT}] \\
    V_{out} & = [\sigma l_{U}(V_{U}) \oplus \sigma l_{PP}(V_{PP}) \oplus \sigma l_{ACT}(V_{ACT})]
    \end{aligned}
\end{equation}
where $\oplus$ represents concatenation, $l_{*}$ represents the linear layer, and $\sigma$ represents a non-linear activation function (in practice, LeakyReLU is adopted).

\smallskip\noindent\textbf{Hypergraph Layer:} is introduced to address the property that each hyperedge can connect more than two nodes. Each Hypergraph Layer is a sequence of a HyperConv~\cite{bai2021hypergraph} sub-layer, an activation sub-layer $\sigma$, and a dropout sub-layer $\beta$. 
\begin{equation}
\small
\begin{aligned}
V_{hyperConv} &= D_{V}^{-1}G_{H}G_{W}D_{E}^{-1}G_{H}^TV_{in}\Theta \\
 V_{out} &= \beta \sigma V_{hyperConv}
\end{aligned}
\end{equation}
where $\Theta$ is the learnable weight matrix in the convolutions.

\smallskip\noindent\textbf{CHAR Layer:} Given a graph's learned node representations $V_{rep} \in R^{(|U|+|PP|+|ACT|) \times d_v}$ and new coming input data $X \in R^{n \times d_x}$ as candidate hyperedges, the CHAR layer tries to predict the connecting nodes for each hyperedge $x_i, i=1, 2, ..., n$ within $X$. First, $X$ is projected into the same dimension as $V$ and the result is activated using non-linear transformations. Then matrix multiplication is performed between the input data and the learned node representation embeddings. The final outputs are the node predictions for each hyperedge.
\begin{equation}
\small
    \begin{aligned}
    X_{PP} &= \sigma l_{PP}(X), X_{PP} \in R^{|PP| \times d_v} \\
    X_{ACT} &= \sigma l_{ACT}(X), X_{ACT} \in R^{|ACT| \times d_v}
    \end{aligned}
\end{equation}
\begin{equation}
\small
    \begin{aligned}
    \hat{Y}_{PP} & = (X_{PP} \otimes V_{PP}^T) \\
    \hat{Y}_{ACT} & = (X_{ACT} \otimes V_{ACT}^T)
    \end{aligned}
\end{equation}
where $\otimes$ represents the matrix multiplication operation.

\smallskip\noindent\textbf{Putting it all together:} Essentially, a GCN~\cite{kipf2016semi} can only capture  the non-heterogeneous and non-hypergraph graphs leading to an inevitable loss of  information~\cite{yang2019revisiting}. Despite the existence of prior work that explicitly addresses hyperedges~\cite{bai2021hypergraph}, the issue of heterogeneity within CHAR data has not previously been considered in a GNN. Thus, our proposed approach innovatively combines the Heterogeneity Layer and the Hypergraph Layer to achieve superior performance. Finally, the CHAR layer is used to predict $\hat{Y}$. In a supervised fashion, the model learns node representations and also predicts node connections giving hyperedges simultaneously.

\section{Experiments}
\subsection{Experiment Dataset}
\label{sec:dataset}
We evaluated \model~on the publicly available Extrasensory~\cite{vaizman2017recognizing} dataset, which contains 6,355,350 instances gathered from 60 participants. Each instance contains 170 features extracted from smartphone sensor data including accelerometer, gyroscope, location, phone state, audio and gravity sensors. 17 \textit{Extrasensory} labels were considered (4 phone placement and 13 activity labels).

\subsubsection{Dataset Pre-processing}

 The original dataset contained signals with varied durations that we sampled using a window length = 3 seconds and step size = 1.5 seconds, the values we experimentally found to be optimal. 
%
\textit{Rules to fix labeling issues: }As a phone can only be carried in one position at a time, phone placement labels are mutually exclusive. Only phone placement labels of $0$, $1$, or $missing$ are valid. In a pre-processing step, intuitive rules were applied to resolve ambiguous and duplicate labels, and remove conflicting activity and phone placement labels such as those that cannot co-occur.

\subsection{Feature Extraction} Handcrafted features commonly utilized in CHAR~\cite{vaizman2018context,ge2022qcruft} were extracted. Features were then normalized using $z = (x-\mu)/s$ where  $\mu$ and $s$ are the mean and standard deviation of features in the training set, before being applied to the validation and testing sets. $x$ and $z$ are the original and transformed features, respectively. 
\vspace{-5pt}
\subsection{Baseline CHAR Models}
We evaluated baseline models including ExtraMLP~\cite{vaizman2018context}, CRUFT~\cite{ge2020cruft}, GCN~\cite{kipf2016semi} and LightGBM~\cite{ke2017lightgbm} as well as our {\model} and its variants: 

\begin{itemize}
    \item \textbf{\model}: proposed approach using optimal hyperparameter values. 
    \item \textbf{Hetero GCN}: HHGNN without hyperedges, to evaluate the contribution of hyperedges to \model 
          performance. 
    \item \textbf{Hyper GCN}: HHGNN without the heterogeneity layer to evaluate the contribution of the heterogeneity layer.
    \item \textbf{1-layer-HHGNN}: demonstrates that using two HHGNN layers in {\model} is optimal. 
\end{itemize}

\vspace{-5pt}
\subsection{Evaluation Metrics}
Considering that the CHAR dataset was extremely imbalanced, the two main evaluation metrics selected were  Matthews Correlation Coefficient (MCC) and Macro F1 Score.

\vspace{-5pt}
\subsection{Experimental Setting}

The dataset was randomly split into 60\% for training, 20\% for validation, and 20\% for hold-out testing. {\model} was trained on the training set, and optimal hyperparameters were selected using grid search based on the best-performing model on the validation set. Evaluation results were reported on the hold-out testing set. The weighted binary cross entropy loss function was adopted as our target loss function (Eq.~\ref{eq:loss}), where the instance-wise pair weights $\omega_{n,c}$ is inversely proportional to the label frequency and set to 0 if the label is missing. The weight ensures that positive and negative instances contribute equally to the loss. For graph representations, $V$ and $G_{attr}$ were initialized to the aggregated mean values of features and were fine-tuned during model training, while other matrices were set to initialization values. All other parameters were randomly initialized with a fixed random seed to enhance reproducibility. The performance of the proposed {\model} was evaluated and reported on three levels: 1) Results for each label, 2) Average results for the activity and phone placement categories, and 3) Average of all activities and phone placement labels.
\begin{equation}
\small
\resizebox{0.9\hsize}{!}{$
    Loss = \frac{1}{N}\sum_{n=1}^N\sum_{c=1}^C[-\omega_{n,c}(y_{n,c}log\hat{y_{n,c}} +(1-y_{n,c})log(1-\hat{y_{n,c}}))]
$}
    \label{eq:loss}
\end{equation}

\begin{table}[t]
    \caption{Major Results on {\model} and baselines.}
    \vspace{-5pt}
	\centering
    \small
    \scalebox{.8}{
	\begin{tabular}{|c|c|c|c|c|c|c|c|}\hline
		 & \multicolumn{2}{c|}{Phone Placement} & \multicolumn{2}{c|}{Activity} & \multicolumn{2}{c|}{Overall} \\ \hline
		 Models & MCC & MacF1 & MCC & MacF1 & MCC & MacF1\\ \hline
         
         ExtraMLP &   0.784   &   0.813   &   0.636   &   0.608   &   0.673   &   0.659\\ 
         CRUFT  & 0.758    &  0.868    &  0.616   &  0.783   &  0.651    & 0.804   \\
		 GCN &   0.824   &   0.907   &   0.675   &   0.813   &   0.712   &   0.837\\
         LightGBM  &   \textit{0.835}   &   \textit{0.913}   &   \textit{0.710}   &   \textit{0.837}   &   \textit{0.741}  &   \textit{0.856}\\
         {\model} &   \textbf{0.953}   &   \textbf{0.976}   &   \textbf{0.808}   &   \textbf{0.896}   &   \textbf{0.845}   &   \textbf{0.916}\\ \hline
		 
	\end{tabular}
	}
	\vspace{-10pt}
	\label{tab:expResultMajor}\\
\end{table}

\begin{table}[t]
    \caption{Ablation Study on {\model}.}
    \vspace{-5pt}
	\centering
    \small
    \scalebox{.8}{
	\begin{tabular}{|c|c|c|c|c|c|c|}\hline
		 & \multicolumn{2}{c|}{Phone Placement} & \multicolumn{2}{c|}{Activity} & \multicolumn{2}{c|}{Overall} \\ \hline
		 Models & MCC & MacF1 & MCC & MacF1 & MCC & MacF1\\ \hline
		 
		 Hetero GCN & 0.916   &   0.957   &   0.754   &   0.863   &   0.795   &   0.887 \\ 
		 Hyper GCN &  0.705   &   0.836   &   0.694   &   0.826   &   0.697   &   0.828 \\
        1-layer-HHGNN &   0.942   &   0.970   &   0.804   &   0.894   &   0.839   &   0.913\\
		 {\model} &   0.953   &   0.976   &   0.808   &   0.896   &   0.845   &   0.916\\ \hline
		 
	\end{tabular}
	}
	\vspace{-10pt}
	\label{tab:expResultAblate}
\end{table}
\vspace{-5pt}
\subsection{Experiment Results}
Results on \textit{Extrasensory} is reported in Table~\ref{tab:expResultMajor} and Table~\ref{tab:expResultAblate}.

\smallskip\noindent\textbf{Overall Performance:} In Table~\ref{tab:expResultMajor}, results for the best performing baseline model under each column is shown in \textit{italic}. If the result of {\model} is significantly better than the best baseline result (gap larger than 0.01), it is marked in \textbf{bold}. 
{\model} outperformed the best baselines with 14.04\% improvement on MCC and 7.01\% improvement on MacF1 over the \textit{Extrasensory} dataset. 

\smallskip\noindent\textbf{Performance in each category:} There were also consistent improvements for {\model} for both Phone Placement and Activity categories in Table~\ref{tab:expResultMajor}. {\model} had a 14.13\% MCC improvement and 6.90\% MacF1 improvement for predicting Phone Placement, and a 13.80\% MCC improvement and 7.05\% MacF1 improvement for predicting Activity, against the best performing baseline LightGBM.

\smallskip\noindent\textbf{Ablation Study:} In Table~\ref{tab:expResultAblate}, when comparing {\model} to its variants, the heterogeneous design and hyperedges contributed non-trivially. When the hypergraph is converted into a traditional graph (i.e., Hetero GCN) by breaking hyperedges into multiple pair-wise edges, the performance of~\model~degraded due to information loss~\cite{yang2019revisiting}. The inferior performance of Hyper GCN compared to {\model} also confirms that node embeddings learned in a unified space do not adequately capture heterogeneity characteristics of context-aware human activity data~\cite{yang2020lbsn2vec++}. Admittedly, the gap between {\model} and the Hyper GCN is larger, indicating that Heterogeneity contributes to performance more than hyperedges. However, this may change in a larger, real-world dataset. Lastly, \model's performance decreased after removing a HHGNN layer. 
Intuitively, two HHGNN layers capture information in two hops of the graph, while one-layer HHGNN only learns representations from one-hop neighbor nodes. The network likely benefited from a two-layer design due to the complexity of in-the-wild CHAR data.

\label{sec:analysis}

\vspace{-5pt}
\section{Conclusion}
To improve on prior approaches to the challenging Context-aware Human Activity Recognition (CHAR) task, we propose a novel graph learning approach, transforming original data-label connections into a Heterogeneous Hypergraph that explicitly encodes previously implicit relationships between context-aware activity labels. Consequently, the multi-label CHAR problem is transformed into a node identification problem given an unknown hyperedge. Transforming real-world datasets yields a graph with heterogenous nodes and hyperedges. {\model}, a novel heterogeneous hypergraph deep learning model is proposed for resolving the newly transformed problem. To the best of our knowledge, ours is the first effort that formulates the CHAR task as a heterogenous hypergraph problem without requiring external information. In rigorous experiments, our proposed approach outperformed SOTA baselines by 14.04\% on Matthews Correlation Coefficient (MCC) and 7.01\% on Macro F1 scores.

In future, we will evaluate {\model} using subject-level splitting and explore other backbone GNN models.
\vspace{-5pt}
\section*{Acknowledgment}
DARPA grant HR00111780032-WASH-FP-031 and NSF grant CNS-1755536 supported this research. 

\bibliographystyle{IEEEtran}
\bibliography{refs}
\flushend


\end{document}